\documentclass[journal]{IEEEtai}
\usepackage{color,array}
\usepackage[english]{babel}
\usepackage{inputenc}
\usepackage{algorithm}
\usepackage{algorithmicx}
\usepackage{algcompatible}
\usepackage{algpseudocode}
\usepackage{makecell}
\usepackage{cite}
\usepackage{pbox}
\usepackage{amsmath,amssymb}
\newcommand\norm[1]{\left\lVert#1\right\rVert}
\usepackage[pdftex]{graphicx}
\usepackage[figurename=Fig.,font=small,skip=2 pt]{caption}
\graphicspath{{../pdf/}{../jpeg/}}
\graphicspath{ {./img/}}
\graphicspath{ {./img1/}}
\DeclareGraphicsExtensions{.pdf,.jpeg,.png}
\usepackage{epstopdf}
\usepackage{multirow}
\usepackage{accents}
\newcommand{\justified}{%
  \rightskip\z@skip%
  \leftskip\z@skip}
\usepackage{multicol}
\usepackage{enumerate}
\usepackage{array}
\usepackage{color,soul}
\usepackage{pbox}

\usepackage{comment}
\usepackage{stfloats}
\usepackage{amssymb}

\usepackage{setspace}
\usepackage{amsmath}
\usepackage{float}
\usepackage[table]{xcolor}
\definecolor{Gray}{gray}{0.85}
\usepackage{verbatim} 
\usepackage{color, colortbl}
\definecolor{Gray}{gray}{0.92}
\def\l{\left}   % left
\def\r{\right}   % Right
\def\defn{\stackrel{\Delta}{=}}

\include{macros}

\begin{document}
% \linenumbers
% \title{Intelligent Fault Diagnosis with Quick Learning Mechanism for Cross Domain Adaptation.}

\title{Quick Learning Mechanism with Cross-Domain Adaptation for Intelligent Fault Diagnosis}

\author{Arun K. Sharma,~\IEEEmembership{Student Member,~IEEE and}
        Nishchal K. Verma,~\IEEEmembership{Senior Member,~IEEE}
      
\thanks{Arun K. Sharma and Nishchal K. Verma are with the Dept. of Electrical Engineering, Indian Institute of Technology, Kanpur, India.
e-mail: arnksh@iitk.ac.in and nishchal@iitk.ac.in}
}

\maketitle
\begin{abstract}
The fault diagnostic model trained for a laboratory case machine fails to perform well on the industrial machines running under variable operating conditions. For every new operating condition of such machines, a new diagnostic model has to be trained which is a time-consuming and uneconomical process.
Therefore, we propose a quick learning mechanism that can transform the existing diagnostic model into a new model suitable for industrial machines operating in different conditions. The proposed method uses the Net2Net transformation followed by a fine-tuning to cancel/minimize the maximum mean discrepancy between the new data and the previous one. The fine-tuning of the model requires a very less amount of labelled target samples and very few iterations of training. Therefore, the proposed method is capable of learning the new target data pattern quickly. The effectiveness of the proposed fault diagnosis method has been demonstrated on the Case Western Reserve University dataset, Intelligent Maintenance Systems bearing dataset, and Paderborn university dataset under the wide variations of the operating conditions. It has been validated that the diagnostic model trained on artificially damaged fault datasets can be used to quickly train another model for a real damage dataset.
\end{abstract}

\begin{IEEEImpStatement}
The operating condition of the real-time machines in the industries may change depending on the loads or applications. For fault diagnosis of such machines, training a diagnostic model for every change in operating condition is not feasible in real-time. 
The quick learning mechanism proposed in this paper solves this problem by transforming the existing diagnostic model from laboratory case machines to real case machines running under any load conditions. The proposed methodology is capable to generate and train a suitable diagnostic model quickly for every new operating condition and therefore this method can be a very promising solution for the real-time monitoring of industrial case machines. 
\end{IEEEImpStatement}

\begin{IEEEkeywords}
Intelligent Fault Diagnosis, Condition Based Maintenance, Transfer Learning, Domain Adaptation, Net2Net operation, Maximum Mean Discrepancy (MMD)  
\end{IEEEkeywords}

\section{\textbf{Introduction}}\label{sec:intro}
\IEEEPARstart{T}{he} continuous monitoring of real-time industrial rotating machines plays a vital role in the safety and productivity of modern industries. It requires a diagnostic model for analysis of the measurement data generated by the continuous processes. For the fault diagnosis of real-time industrial machines, there are two main challenges: (i) unavailability of the labeled dataset as some machines may not be allowed to run in a failure state, which may cause catastrophic accidents or critical breakdown and (ii) distribution of the dataset may change due to change in the operating condition of the machine. 

% With the advent of modern computation resources, intelligent systems for condition-based monitoring have become very popular, \cite{rn2, rn1, rn3}.
% The diagnostic models trained for laboratory machines may not be suitable for the industrial machines due to different operating conditions. Training a deep learning model for the analysis of every new set of data is impractical due to the time complexity of the model, \cite{rn2, rn1, rn3}. 
% Therefore, the analysis of a dataset from such processes requires an intelligent system that can retain the previously learned knowledge and utilize it to quickly learn the new task.
In literature, various data-driven approaches have been reported for intelligent health monitoring of the rotating machines also called condition-based maintenance (CBM). The intelligent systems based on evolutionary algorithm \cite{cRe1, cRe2, cRe3}, fuzzy systems \cite{aks2}, and deep neural networks \cite{sparseAE, cnn} are applied for learning the complex pattern from recorded data. It monitors specific changes in the machine signatures like vibration, acoustics, temperature, pressure, etc, \cite{sr1,sr11, sr3} and notify the anomaly/fault in the various component of the machine.

Vibration-based CBM has gained much attention due to its effectiveness and flexibility in assessing the machine health conditions using time domain and frequency domain methods of signal analysis, \cite{sr2, vib1, vib3, vib4}. But, these approaches become ineffective if there is a contamination of extraneous frequencies and there is a change in the dynamic behaviour of the machine.
The recent advancement in machine learning techniques overcomes these problems and is capable to learn highly non-linear and complex characteristics of a data pattern and diagnose the machine health condition \cite{rn2, rn1, rn3, R5, rzhao, app_ml_fault}. These methods give accurate/commendable results for the test dataset exactly similar to the training data. 
If the operating condition of the machine is variable, the diagnostic model fails to perform well unless the model is retrained for the new data pattern. For such problems, knowledge transfer (also, called transfer learning) gives a better solution, \cite{pratt, sjPan, R.kula_bio}, where training is accelerated by transferring the previous knowledge to the new similar task.

LY Pratt in 1993, \cite{pratt} formulated a discriminability-based transfer algorithm to introduce the concept of transfer learning. It uses previously learned knowledge to initialize the parameter of the target model to reduce the time of training.
In most of the approaches of the transfer learning \cite{sjPan}, the pre-trained model on the source dataset is fine-tuned using a similar target dataset. If the distribution of the target data is different, fine-tuning the model requires a large amount of labeled target data and a long training process. For such cases, the transfer learning method fails to train the model using a small number of training samples from the target domain.  To solve the problem of cross-domain fault analysis, domain adaptation techniques have been reported in various literature \cite{sjPan2011, longRTRL, dann_jmlr, lwen, wLu, cross-domain, dctln}.

S.J. Pan \textit{et. al.} \cite{sjPan2011} proposed a learning mechanism for shared subspace using transfer component analysis via minimizing maximum mean discrepancy (MMD) of the source and the target features representations. Y. Ganin \textit{et. al.} \cite{dann_jmlr} proposed a domain adversarial training of DNN which uses labeled source data and unlabeled target data. L. Wen \textit{et. al.} \cite{lwen} has suggested the minimization of combined objective function of classifier loss and the MMD term for the domain adaptation during fine-tuning of the model. Similarly, W. Lu \textit{et. al.} \cite{wLu} uses MMD term to model the loss due to change in probability distribution in a subspace and obtained a training law for DNN. 
These methods of transferring knowledge are not useful if the change in the data pattern needs a deep learning model with different architecture.

% G. wang \textit{et al}, \cite{tl-gdbn} proposed a growing deep belief network that automatically develops its optimal architecture using the concept of knowledge transfer from the learned neurons to the newly added neuron nodes. This gives a promising solution to get an optimal network architecture for a specific application. But, it becomes difficult if there is a need to quickly change the architecture for the variable working condition. 
T. Chen \textit{et al}, \cite{Ian} has proposed Net2Net transformation to initialize a new (student) network from a previously trained (teacher) network based on the function preserving principle. The function preserving principle allows us to quickly change the architecture of the network without changing the function mapping. If the student network with new architecture has to be used with a new dataset of a different probability distribution, it requires a sufficient amount of data and a large number of iterations to learn the new data pattern. Again, it becomes a challenge to quickly adapt with the change in the data pattern during the continuous process of experimentation.

To solve this problem, we propose a quick learning mechanism based on Net2Net transformation followed by fine-tuning with a domain adaptation algorithm. The process of fine-tuning minimizes the classification loss plus domain discrepancy using a few number of samples from the target domain. The key contributions of this work are summarized as follows:

\begin{enumerate}[i)]
\item The proposed mechanism is capable to train a DNN of user-provided architecture needed for the fault diagnosis of machines under variable working conditions. 
\item The main novelty of the proposed work is that the desired architecture suitable for the industrial machine is generated from a trained DNN. The fine-tuning of the new DNN includes minimization of classification loss as well as MMD term. Therefore, the new DNN requires a less amount of target data and a few iterations of training. This makes the algorithm to learn the new pattern of the dataset faster.
\end{enumerate}

The rest of the paper is organized as follows: Section \ref{sec:dnn} briefly describes the theoretical background of DNN, domain adaptation, and function preserving network transformation.
Section \ref{net2net} describes the development of quick learning system: the proposed approach.  In section \ref{sec:results}, effectiveness of the proposed fault diagnosis approach has been demonstrated on : (i) Case Western Reserve University (CWRU) fault diagnosis bearing data \cite{cwru}, (ii) Intelligent Maintenance Systems (IMS) bearing dataset \cite{imsnasa}, and Paderborn university dataset \cite{paderborn}. Finally, section \ref{sec:conclusions} concludes the paper.
%###############################################################################

\section{Theoretical Background}\label{sec:dnn} 
\subsection{Deep Neural Network} 
\begin{figure}[!h]
\centering
\includegraphics[width=8.5cm]{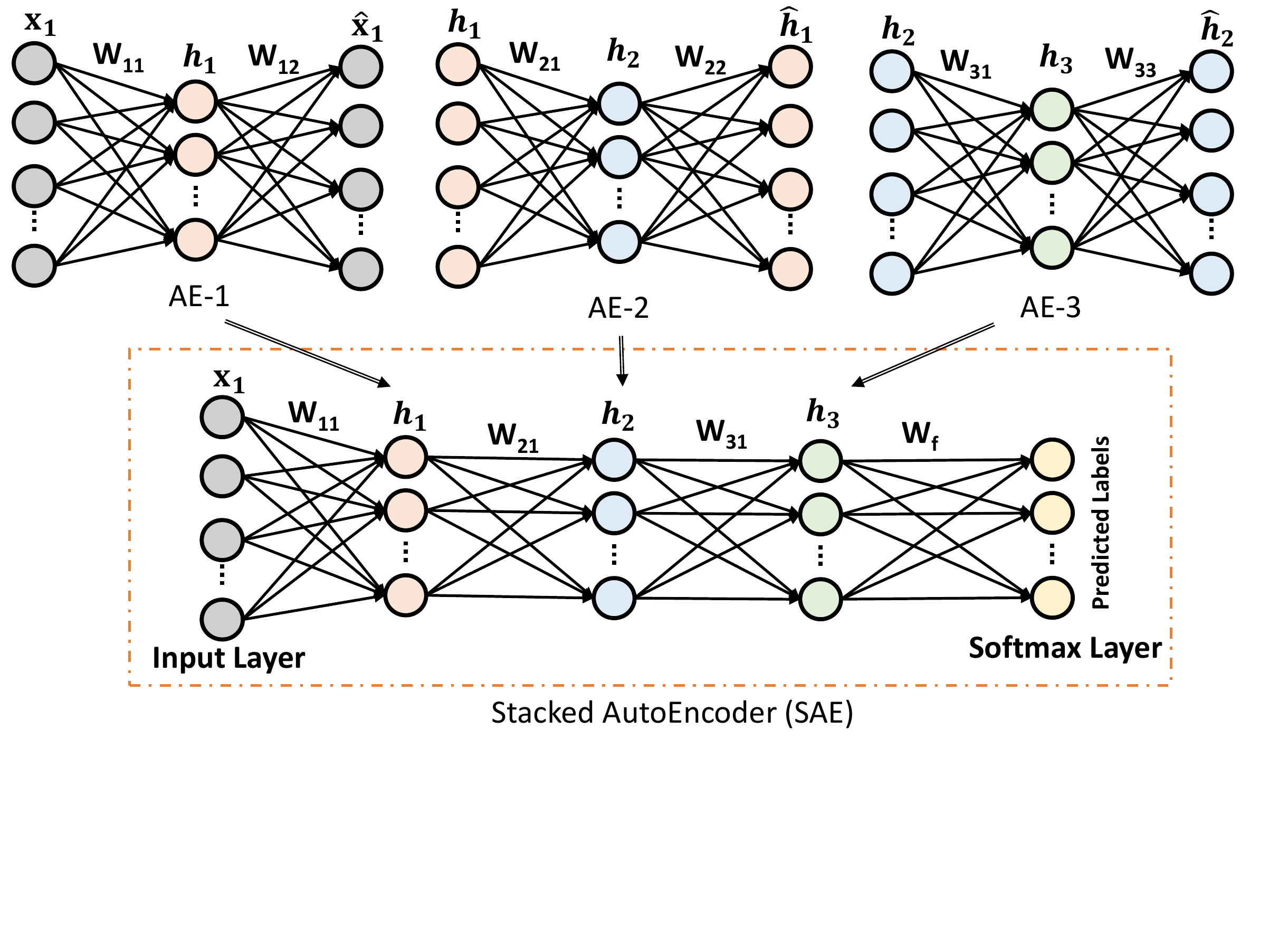}
\vspace{-1.3cm}
\caption{Formation of stacked auto-encoder with three hidden layer}
\label{fig:SAE}
\end{figure}

Deep neural network (DNN) is a multi-layer neural network capable of highly non-linear transformation through each layer depending on the types of activation function chosen to fit the problem. The training of the network includes obtaining optimal weight to map the input-output dataset.
The DNN discussed in this section is a multi-layered network called stacked auto-encoder (SAE) \cite{bengio}, which is trained by the method of greedy layer unsupervised learning using unlabeled data followed by fine-tuning using labeled data. The formation of SAE by stacking auto-encoders is depicted in Fig. \ref{fig:SAE}.

\textit{Learning Mechanism of Autoencoder:}
The autoencoder is trained to learn the identity approximation \cite{bengio}, of the unlabeled dataset, $h_{W,b}(X)\approx X$, where $W$ denotes the weight matrix and $b$ be the bias vector of the hidden layer of the autoencoder.
For an unlabeled dataset, the cost function $J$ for sparse autoencoder in term of weight $W$ and bias $b$ is defined as

\begin{align}\label{cost:sae}
J = \frac{1}{N}\sum_{i=1}^N\frac{1}{2}\norm{h_{\textbf{W,b}}(\textbf{X}^l)-\textbf{X}^l}^2 + \frac{\lambda}{2}\sum_{i=1}^{d_l}\sum_{j=1}^{d_{l+1}}(W_{ji})^2\nonumber\\ + \beta\sum_{j=1}^{d_l}{KL}(\rho/\hat{\rho})
\end{align}
where, $N$ be the number of training samples, $\textbf{X}^l$ be the input samples to $l^{th}$ autoencoder, $\lambda$ be the regularization parameter, $d_l$ denotes the number of nodes in the $l^{th}$ layer, $\rho$ be the sparsity parameter, $\beta$ be the weight penalizing deviation from $\rho$ and $KL(.)$ denotes the Kullback-Leibler divergence function. 
Sparsity is the mean of activation of a hidden layer averaged over the training set and is enforced to be equal to a given sparsity parameter. Therefore, sparsity constraint ensures the desired sparsity of the generated feature representation(s) at the hidden layer. Regularization parameter \cite{regularization} is used to ensure the appropriate fitting of the DNN hyper-parameters for the training dataset and avoid over-fitting or under-fitting for the testing on unseen data.

\subsection{Domain Adaptation}\label{DoAdapt} Domain adaptation in transfer learning creates a self-taught system that learns the target data-mapping without the availability of labeled data or with partial availability of labeled datasets. Domain adaptation by minimizing maximum mean discrepancy has gained much popularity due to its capability of domain shift in the shared subspace.

\subsubsection{Maximum Mean Discrepancy (MMD)} MMD gives the measure of non-parametric distance between mean of two distributions on reproducing kernel Hilbert space (RKHS) \cite{mmd, rkhs}.
% It estimates non-parametric distance between two distributions on the space of probability measure as the distance between the corresponding two mean elements. 
For a universal RKHS, MMD asymptotically becomes zero if and only if the probability distribution of the two dataset are same. Let $x_i^s\in \textbf{X}^s$ and $x_j^t\in \textbf{X}^t$ be the $i^{th}$ source data and the $j^{th}$ target data in the source space $\chi^s$ and the target space $\chi^t$ respectively, then the MMD of the two distribution in RKHS is defined as
\begin{equation} \label{mmd}
{\rm {MMD}}(\textbf{X}^s,\textbf{X}^t) = \left\Vert {\frac{1}{{n^s}}\sum \limits _{i = 1}^{{n^s}} {f \left({{x_i^s}} \right)} - \frac{1}{{n^t}}\sum \limits _{j = 1}^{{n^t}} {f \left({{x_j^t}} \right)} } \right\Vert _{\mathcal H} 
\end{equation}
where, $f(.)$ denotes the kernel function $f:X,X\to {\mathcal H}$, ${\mathcal H}$ be the universal RKHS, $n^s$ and $n^t$ be the number of samples in the source and the target datasets $\textbf{X}^s$ and $\textbf{X}^t$ respectively. 

\subsubsection{Function Preserving Transformation} 
A function $y=f(x;\psi)$ represented by an neural network (NN) model can equivalently be represented by another NN model $g(x;\psi')$ if a new set of parameter $\psi'$ is chosen such that 
\begin{align}
    \forall x, \;\;f(x;\psi) = g(x;\psi')
\end{align}
where, $x$ is the input dataset, $y$ is the output class label, $\psi$ and $\psi'$ are the NN parameter vector. This concept allows us for two types of network transformations: {Net2WiderNet} and {Net2DeeperNet}.
% This theorem states that if any change is made to the network after initialization, it is guaranteed to be an improvement in the network convergence provided that the each local step is an improvement.

\subsubsection{Net2WiderNet} \label{sub:wider}
\begin{figure}[!h]
\centering
\includegraphics[width=8.5cm]{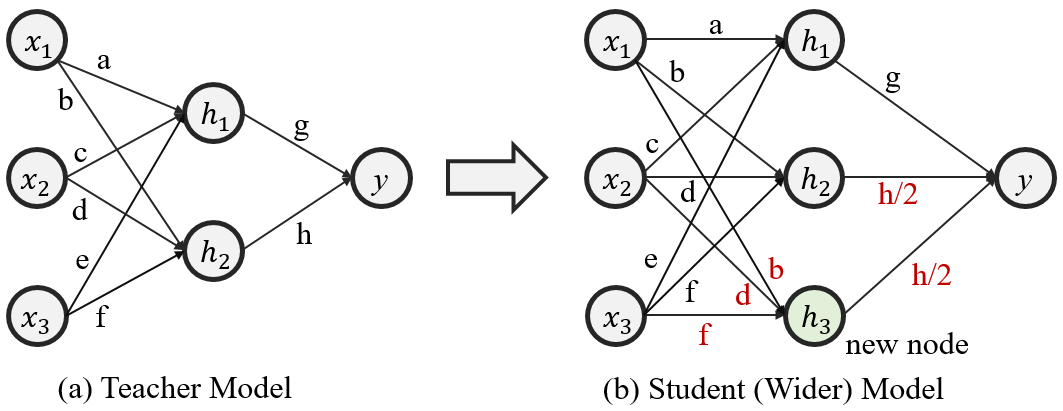}
\caption{Net2WiderNet transformation: $h_3$ be the new node}
\label{fig:widernet}
\end{figure}

Net2WiderNet transformation is used to widen the network by adding nodes in any of the hidden layers of a network, \cite{Ian}. We consider a simple example to elaborate on the idea of replacing the network with a wider network as shown in Fig.\ref{fig:widernet}.

Let the network in Fig. \ref{fig:widernet}(a) is trained on a given dataset and is treated as the teacher network to train a wider network, called student network in Fig. \ref{fig:widernet}(b). The knowledge of the network in Fig. \ref{fig:widernet}(a) is transferred to initialize the parameters of a wider network in Fig. \ref{fig:widernet}(b) such that it provides the same output as the network in Fig. \ref{fig:widernet}(a). Random nodes are picked from the hidden layer and replicated as new nodes. Here, the weights of the $h_2$ node ($b$, $d$ and $f$) are replicated as the connections to the new node $h_3$. The outgoing weights are divided by $2$ to compensate for the replication of $h_2$.

The above idea can easily be generalised with a recursive function for more than one layer and to add more than one node in a layer. 
% Assume that weights associated with the layers ${l-1,l}$ and ${l+1}$ of a trained network are $W_{l ,l-1}\in \mathbb {R} ^{n2\times n1}$ and $W_{l+1 ,l}\in \mathbb {R} ^{n3\times n2}$, where,  ${n_1}, {n_2}$ and ${n_3}$ number of nodes in the layers ${l-1,l}$ and ${l+1}$ respectively.
Let us assume that a trained network has $n_2$ number of nodes in the $l^{th}$ layer and has to be widen to $n'_2 (>n_2)$. To initialize the weight matrix in the new network, a random sampling function $\hbar:\{1,2,\cdots, n'_2\} \rightarrow \{1,2,\cdots,n_2\}$ is defined for the repetition of nodes as.
\begin{align} \label{eq:rand_mapping}
\hbar(j) \defn
\left\{
  \begin{array}{ll}
  j  &      j \leq   n_2\\
  \text{random sample from}\; \{1, \dots,n_2\} &  j > n_2
  \end{array}
\right.
\end{align}
where, $j$ is the node index from $1$ to $n'_2$ in the new network. Once, random repetition of nodes are chosen, weights are assigned as demonstrated in Fig. \ref{fig:widernet}.

\subsubsection{Net2DeeperNet} \label{sub:deeper} 
{Net2DeeperNet} transformation allows us to insert a new layer to the network to transform it into a deeper network \cite{Ian}.
\begin{figure}[h!] 
\centering
\includegraphics[width=8.5cm]{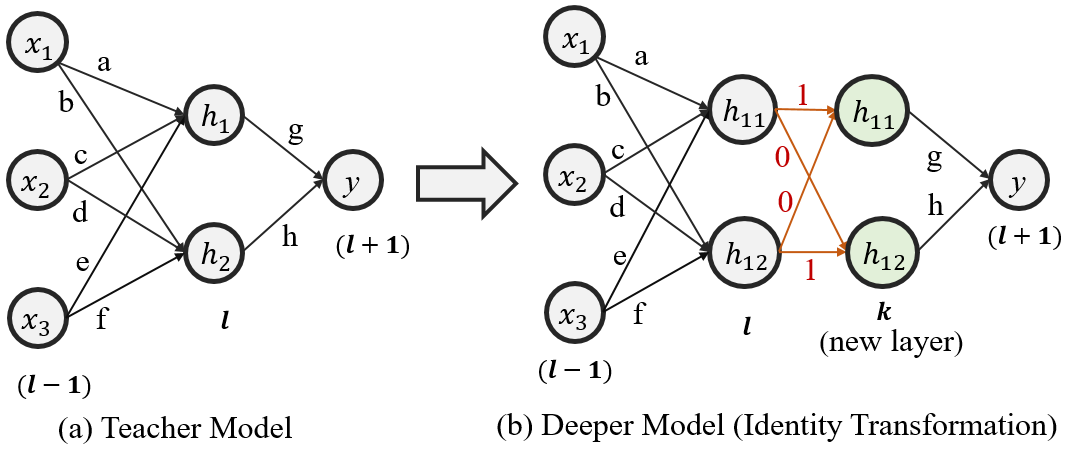}
\caption{Net2DeeperNet transformation: $k^{th}$ be the new layer}
\label{fig:archi_deeper}
\end{figure}

In Fig. \ref{fig:archi_deeper}, deeper model is created by inserting a layer $k$ after the hidden layer $l$ having same number of nodes as in the layer $l$. Let the output of the $l^{th}$ layer be $v^{(l)} = \phi(v^{(l-1)T}W^{(l)})$
% \begin{align}\label{addLayer1}
%     v^{(l)} = \phi(v^{(l-1)T}W^{(l)})
% \end{align}
where, $\phi(.)$ represents the mapping by the $l^{th}$ layer and $W^{(l)}$ be the weight matrix in between $(l-1)^{th}$ and $l^{th}$ layers. The new $k^{th}$ layer is inserted such that its output is is given by
\begin{align}\label{addLayer2}
    v^{(k)} = \phi\l(W^{(k)T}\phi\l(v^{(l-1)T}W^{(l)}\r)\r)
\end{align}
where $W^{(k)}$ is the weight matrix in between $l^{th}$ and $k^{th}$ layers and initialized to be identity matrix.

\section{Quick Learning Mechanism: Proposed Methodology}
\label{net2net} We propose a network transformation method using the function preserving principle of the NN model and minimization of MMD term from source to target dataset. 
% First, a suitable model for the new dataset, called student network is initialized by the knowledge contained in the teacher network. Then, the network with a new architecture is fine-tuned with stochastic gradient descent along with weight shifting to minimize the MMD of h-level output for the source and the target dataset. 
The flowchart of the proposed method is depicted in Fig. \ref{fig:net2net}. For the process shown in Fig. \ref{fig:net2net}, ${\rm X^{(s)}}$ denotes the source dataset, ${\rm X^{(t)}}$ the target dataset, ${\rm W}_{te}, \, {\rm b}_{te}$ denotes the weight and bias matrices of the teacher network and ${\rm W}_{stu}, \,{\rm b}_{stu}$ of student network. The whole learning mechanism includes two steps as described in the following subsections:
\begin{figure}[!ht]
\centering
\includegraphics[width=8.8cm]{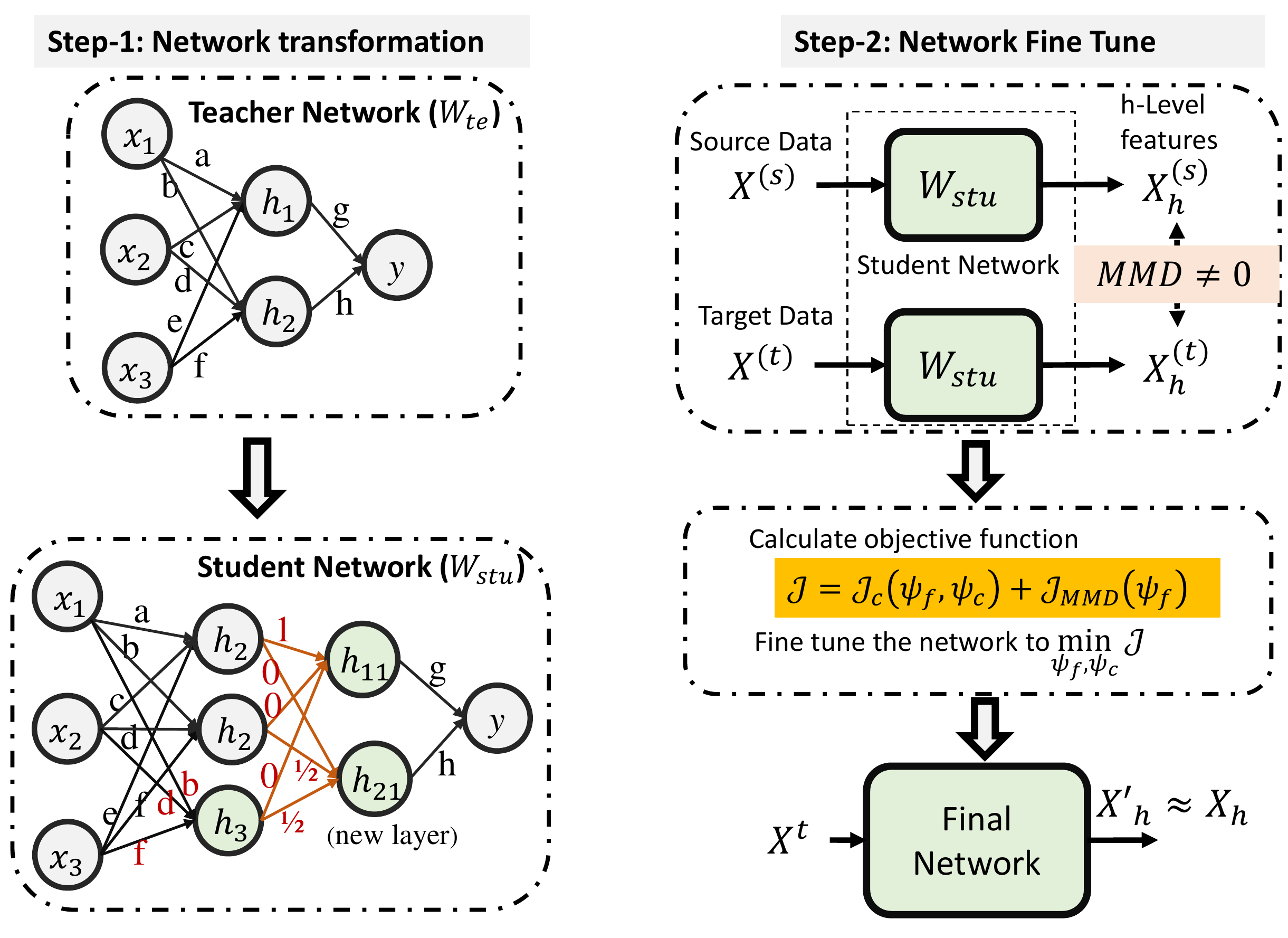}
\vspace{-0.4cm}
\caption{Flowchart of the quick learning methodology (Width and depth of the DNN shown here are for demonstration only)}
\label{fig:net2net}
\end{figure}

%==========================================================
%**********************************************************
\begin{algorithm*}
\caption{Train the Teacher Network (DNN Model)}
\label{algo:Te_DNN}
\begin{algorithmic}[1]
\State{Train Auto-Encoder (AE)}
\For{ i $\in$ Number of hidden layers}

     $AE_i \longleftarrow [h_{i-1},\, h_{i}, \,h_{i-1}]$ \hspace{10pt} \# Train the $i^{th}$ AE, $h_i=i^{th}$ hidden layer  $i-1$ for $i=1$ indicates the input data \EndFor
\item $\textrm{DNN} \longleftarrow [\textrm{Input},\, h_{1},\, h_{2}, ..., h_{l}, \textrm{Softmax layer}]$\hspace{10pt} \# Form DNN by stacking all hidden layers with input and output layer
\item Classification loss $J(W, b) \longleftarrow$ MSE [true label, predicted label]  \hspace{15pt}  \#  Calculate Mean Square Error (MSE)
\item Fine tune the network to find the optimal parameter (weight and bias vectors)
\item Teacher network $\longleftarrow$ Train DNN
\end{algorithmic}
\end{algorithm*}

%%################################################################
\begin{algorithm}[!ht]
\caption{Algorithmic Steps for Quick Learning Mechanism: The Proposed Methodology }\label{algo:net2net}
\begin{algorithmic}[1]
\State\textbf{begin}
\State{\textbf{Inputs}: } Teacher Network (${\rm W}_{te}$, ${\rm b}_{te}$) and target datasets
\For{ l $\in$ layers to be replicated}
    \begin{enumerate}[a)]
       \item Add new layer $k$ after layer $l$ of size same as $l$
       \item Assign weight matrix to new layer $U^{(l)} = I_n$ (Identity matrix) to satisfy (\ref{addLayer2})
        % \item Keep weight of all other layers unchanged.
    \end{enumerate}
\EndFor 
\For{l $\in$ layers to be widened}
\begin{enumerate}[a)]
      \item Let number of nodes in layers ${l-1,l}$ and ${l+1}$ be  ${n_{l-1}}, {n_l}$ \& ${n_{l+1}}$ and number of nodes to be added = $n$. 
      \item Generate random samples $\hbar\in [1, n_l]$ using equation (\ref{eq:rand_mapping})
      \item Let $r_j$ denotes the number of repetition of $j^{th}$ node in random samples $\hbar$
      \item Let ${\rm W}^{(l)}_{j,i}$ denotes the weight connecting $i^{th}$ node in $(l-1)^{th}$ layer to $j^{th}$ node in $l^{th}$ layer
      \item ${\rm W}^{(l+1)}_{k,j}$ denotes the weight connecting $j^{th}$ node in $l^{th}$ layer to $k^{th}$ node in $(l+1)^{th}$ layer
      \For{${j} \in [n_l+1,n_l + n]$}
        \begin{enumerate}[]
            \For{$i\in [1,n_{l-1}]$}
                    $${\hspace{1cm}\rm W}^{(l)}_{j,i} = {\rm W}^{(l)}_{\hbar_{j-n_l},i} \;\;\& \;\;{\rm b}^{(l)}_{j} \;=\; {\rm b}^{(l)}_{\hbar_{j-n_l}} $$
            \EndFor
        \end{enumerate}
        \begin{enumerate}[]
            \For{$k\in [1,n_{l+1}]$}
                $$\textrm{Assign } {\rm W}^{(l+1)}_{k,j} \;=\; {\rm W}^{(l+1)}_{k,\hbar_{j-n_l}}/r_j\;$$
                $$\textrm{Update }{\rm W}^{(k)}_{k,\hbar_{j-n_l}} = {\rm W}^{(l+1)}_{k,\hbar_{j-n_l}}/r_j$$
            \EndFor
        \end{enumerate}
      \EndFor
\end{enumerate}
\EndFor
\State Result: student network (weight matrix: $\rm W_{stu}$, ${\rm b}_{stu}$).
\State {Fine-tune the Network:} Update the weight and bias using equations (\ref{updateW1}) and (\ref{updateW2}) to get the optima weight matrix.
\State\textbf{end}
\end{algorithmic}
\end{algorithm}
%%########################## End of Algorithm ######################################
%***********************************************************
%===========================================================

\subsection{Step-1: Network Transformation}
The knowledge transfer using the function preserving concept allows us to transform the network to a new architecture without losing the function mapping (as described in section \ref{sub:wider} and \ref{sub:deeper}). Let ${\rm W}_{te}, \, {\rm b}_{te}$ be the weight and bias matrices of the teacher network. Using Net2Net transformation initial weight and bias ${\rm W}_{stu}$ and ${\rm b}_{stu}$ of student network are obtained then, the network weights are fine-tuned using the new (target) dataset by incorporating MMD term with the classifier loss.

\subsection{Step-2: Fine-tune with Domain Adaptation}
The classification loss for the $C$ class problem and the MMD term is defined using the h-level feature output of the new architecture for the source dataset and target dataset as follows.
\begin{equation}
  \resizebox{0.33\textheight}{!}{%
  $
{{\mathcal J}_{MMD}} = \sum_{i=1}^{C}{\left\Vert {\frac{1}{N_i^{(s)}}\sum \limits_{p = 1}^{N_i^{(s)}} {f (x_{i,p}^{(s)})} - \frac{1}{N_i^{(t)}}\sum \limits_{q = 1}^{N_i^{(t)}} {f (x_{i,q}^{(t)})}} \right\Vert _{\mathcal H}^2}$}
\end{equation}
\begin{equation}
    \resizebox{0.33\textheight}{!}{%
$\mathcal{J}_{c} = \frac{1}{N^{(s)}}\l[\sum_{p=1}^{N^{s}}\sum_{i=1}^{C}I[y_p = C]\log{\frac{e^{(w_i^Tf(x_p^{s})+b_i)}}{\sum_{i=1}^C e^{(w_i^Tf(x_p^{s})+b_i)}}}\r]$}
\end{equation}

where, $f(x)=\Phi(Wx+b)$ is the h-level features representation of DNN, $N_i^{(s)}$ and $N_i^{(t)}$ are the number of samples in the $i^{th}$ of the source dataset ${\rm X^{(s)}}$ and the target dataset ${\rm X^{(t)}}$ respectively. $N^{(s)}$ be the batch size of the source data, $y_p$ be the source label and $[w_i,\, b_i]$ be the weight and bias connecting $i^{th}$ node in the output (softmax) layer. Parameters of the student network are optimized by minimizing the loss function (\ref{doAdapt}).
\begin{equation}
    \mathcal{J} = \mathcal{J}_{c}  + \lambda\mathcal{J}_{MMD}
    \label{doAdapt}
\end{equation}
Let $\psi_f=[{\rm W_{stu}}, \, {\rm b_{stu}}]$ and $\psi_c = [\rm{ W_c,\, b_c}]$ are the parameters of the feature extractor (DNN) and classifier respectively, then the loss function can be minimized to obtain the optimal network parameters by solving (\ref{Eq:MinArg}).
\begin{align}
	\min_{\psi_f,\, \psi_c} \mathcal{J} =  \min_{\psi_f,\, \psi_c} \left[\mathcal{J}_{c}(\psi_f, \psi_c)  + \lambda\mathcal{J}_{MMD}(\psi_f)\right]
	\label{Eq:MinArg}
\end{align}
The cost function optimization objective stated in the equations (\ref{Eq:MinArg}) is achieved by Limited-Broyden-Fletcher-Goldfarb-Shanno (L-BFGS) \cite{bfgs} algorithm or optimal weights can also be obtained using gradient descent as follows
\begin{align}\label{updateW1}
    {\rm \psi_f} \;\gets  & \; {\rm \psi_f} -\eta\left[ \frac{\partial{\mathcal{J}_{c}(\psi_f, \psi_c)}} {\partial {\rm \psi_f}} + \lambda\frac{\partial{\mathcal{J}_{MMD}(\psi_f)}} {\partial {\rm \psi_f}} \right]\\\label{updateW2}
    {\rm \psi_c} \;\gets &\;{\rm \psi_c} -\eta \frac{\partial{\mathcal{J}_{c}(\psi_f, \psi_c)}} {\partial {\rm \psi_c}}
\end{align}
where, $\eta$ is the learning rate for the parameter update.

The training process in the proposed method requires a pre-trained DNN, therefore, the algorithmic steps for the training of DNN is given in the Algorithm \ref{algo:Te_DNN}. The algorithmic steps for the proposed methodology is presented in Algorithm \ref{algo:net2net}.

\section{Results and Discussion} \label{sec:results}
Effectiveness of the proposed framework of network transformation with domain adaptation has been demonstrated using CWRU fault diagnosis bearing data \cite{cwru}, IMS bearing dataset \cite{imsnasa} and Paderborn university dataset \cite{paderborn}.

\subsection{Dataset Description} 
\subsubsection{CWRU Bearing Data \cite{cwru}} \label{sub:cwru_data}
The bearing dataset provided by CWRU was recorded on a ball bearing testing platform. Using electro-discharge machining, motor bearings were seeded with faults of fault diameters 7, 14, and 21 mils (1 mil = 0.001 inches) at the inner raceway, rolling element (i.e. ball), and outer raceway. 
The vibration data were recorded with motor loads of 0 to 3 HP and motor speeds of 1730 to 1797 RPM in four different cases: (i) \textit{normal baseline data} (ii) \textit{$12k$ samples/sec drive end (DE) fault data}, (ii) \textit{$48k$ samples/sec drive end (DE) bearing fault data} and (iv) Fan-End (FE) bearing fault data (recorded at $12k$ samples/second). The vibration signal represents four different states of the machine: (i) healthy (Normal: N), (ii)  inner race (IR), (iii) outer race (OR) and (iv) rolling element (ball: B).

\subsubsection{IMS Bearing Dataset \cite{imsnasa}}  \label{sub:nasa_data}
IMS bearing datasets \cite{imsnasa} are provided by NSFI/UCRC of Intelligent Maintenance System (IMS). It consists of three datasets recorded using high sensitivity quartz ICP accelerometers installed on four bearing housing (Bearing-1 to 4: two accelerometers at each bearing for dataset-1, one accelerometer at each bearing for datasets-2 and 3). Each dataset contains a vibration signal recorded for $1\; sec.$ at the sampling rate of $20kHz$. 
In the case of dataset-1, bearing-3 and bearing-4 get inner race defect and roller element defect respectively at the end of the test-to-failure experiment. 
Bearing-1 in dataset-2 and bearing-3 in dataset-3 get outer race defect at the end of the test-to-failure.

\subsubsection{Padeborn university dataset \cite{paderborn}} \label{sub:paderborn_data}
The Paderborn university dataset is the best dataset for monitoring of bearing faults of electromagnetic rotating machines under a wide variety of operating conditions. It contains recorded signals of 32 different bearing experiments categorized as:
\begin{enumerate}[i)]
    \item 6 experiments on healthy bearings.
    \item 12 experiments on artificially damages bearings
    \item 14 experiments on real damaged bearings by accelerated lifetime tests.
\end{enumerate}
Dataset from each experiment have measurements of motor phase currents, vibration, speed, torque, bearing temperature and radial force. Each dataset contains 20 measurements of 4 seconds under four different settings of speed, torque and force. The four settings are  (i) \textbf{L1: N09\_M07\_F10} (speed = 900 rpm, torque = 0.7 Nm \& radial force = 1000 N) (ii) \textbf{L2:  N15\_M01\_F10} (speed = 1500 rpm, torque = 0.1 Nm \& radial force = 1000 N) (iii) \textbf{L3: N15\_M07\_F04} (speed = 1500 rpm, torque = 0.7 Nm \& radial force = 400 N) and (iv) \textbf{L4: N15\_M07\_F10} (speed = 1500 rpm, torque = 0.7 Nm \& radial force = 1000 N).  There are two categories of faults \textbf{inner race (IR) damage} and \textbf{outer race (OR) damage}. Each category of faults contains wide variety of damages and different level of damage represented by the extent of damage (details can be found in \cite{paderborn}).

\begin{table}[!ht]
\centering %
\caption{\textsc{  \small CWRU \cite{cwru} and IMS \cite{imsnasa} Dataset Description}} %
\label{tab:dataD1}
\resizebox{0.36\textheight}{!}{%
\begin{tabular}{ |c|c|c|c|c|c|c|c|}
\hline
\multirow{2}{*}{\makecell{Class \\ Name}}  &  \multicolumn{2}{c|} {Source (CWRU-DE)} & \multicolumn{2}{c|} {Target-1 (CWRU-DE)}  & \multicolumn{2}{c|} {Target-2 (IMS Data)} & \multirow{2}{*}{\makecell{Class\\ Label}}\\
\cline{2-7}
  &  \makecell{Sample/\\Class} & Load & \makecell{Sample/\\Class} & Load  &  \makecell{Sample/\\Class} & Load & \\
\hline
 N & 1210  & 0 hp  & 400 & 1, 2 \& 3 hp & 400 &26.6 kN  &0  \\
\hline
 IR & 1210   & 0 hp & 400  & 1, 2 \& 3 hp & 400 & 26.6 kN  &1 \\
\hline
 B & 1210   & 0 hp& 400  & 1, 2 \& 3 hp & 400  & 26.6 kN  & 2 \\
\hline
 OR & 1210   &0 hp & 400  & 1, 2 \& 3 hp &400  & 26.6 kN  & 3 \\
\hline
\end{tabular}}
\end{table}
\begin{table*}[!ht]
\centering %
\caption{\textsc{ Paderborn University Dataset \cite{paderborn} description}} %
\label{tab:dataD2}
\resizebox{0.74\textheight}{!}{%
\begin{tabular}{ |c|c|c|c|c|c|c|c|c|c|c|}
\hline
\multirow{2}{*}{\makecell{Class \\ Name}}  &  \multicolumn{2}{c|} {Source (Artificial Damage)} & \multicolumn{3}{c|} {Target-3 (T3)}  &  \multicolumn{3}{c|} {Target-4 (T4)} & \multirow{2}{*}{\makecell{Class\\label}}\\
\cline{2-9}
  &  \makecell{Bearing name\\ (Extent of damage = 1)} & Sample/Class & Bearing name &\makecell{Extent of \\damage} & Sample/Class &   Bearing name & \makecell{Extent of \\damage} & Sample/Class & \\
\hline
 N & K001+K002 & $10000$  & K001 & None & $500$ &  K002 & None & $500$    &0  \\
\hline
OR & KA01+KA05 & $10000$  & KA04 & 1 & $500$ &  KA16 & 2 & $500$   &1  \\
\hline
IR & KI01+KI05 & $10000$  & KI16 & 3 & $500$ &  KI18 & 2 & $500$   &2  \\
\hline
\end{tabular}}
\end{table*}
\begin{figure*}[!ht]
\centering
\includegraphics[width=18.0cm]{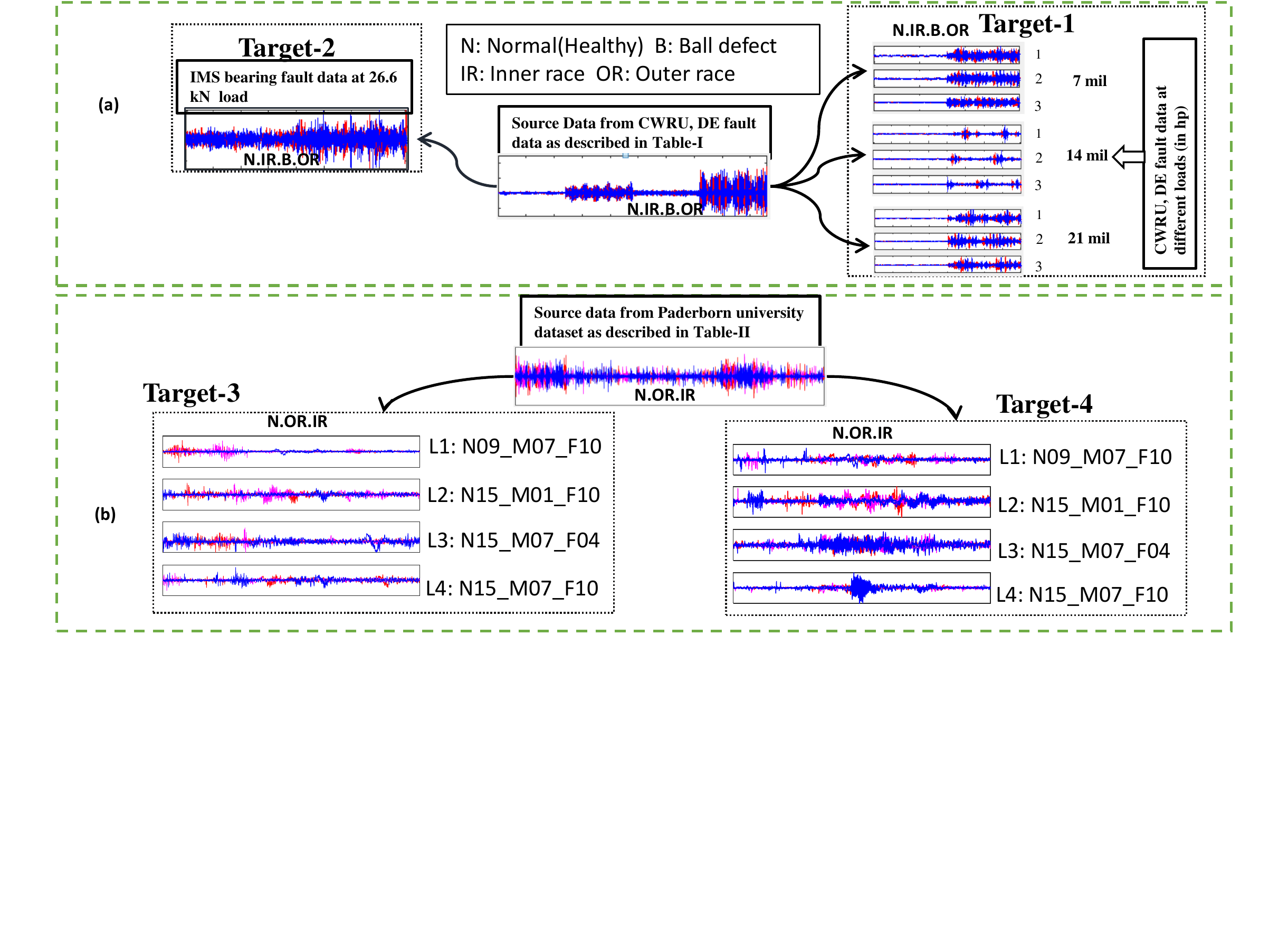}
\vspace{-4.0cm}
\caption{Domain change from source to target for the two cases: (a) Source and target datasets as described in Table \ref{tab:dataD1} and (b) Source and target data as described in Table \ref{tab:dataD2} (The waveform in color red, blue and magenta are the vibration signals).}
\label{Fig: flow diagram}
\end{figure*}

\subsection{Pre-processing} 
The sensor data are usually contaminated with noises and not well structured which makes it unsuitable for the training of a network. Data preprocessing is required which involves filtering, clipping, smoothing and normalization to convert the dataset well structured.
We have used the pre-processed data and re-scaled into [0, 1] using min-max normalization before training the network.
\begin{align}
        \rm {\textbf{x}_{normalized} = \frac{\textbf{x}-\textbf{x}_{min}}{\textbf{x}_{max}-\textbf{x}_{min}}}
\end{align}
where, $\textbf{x}$ is the un-normalized data, $\textbf{x}_{normalized}$ is the normalized data,  $\textbf{x}_{min}$ and $\textbf{x}_{max}$ are minimum and maximum values of the data.  Now, the data is is split into train \& test datasets using five fold cross-validation sampling technique for better generalization of the model.

\begin{table*}[ht]
\centering %
\caption{\textsc{  \small  Accuracy on CWRU DE Fault Dataset and IMS Dataset (Source Data: 0 hp, 7 mil, DE Fault)}} %
\resizebox{0.68\textheight}{!}{%
\begin{tabular}{ |c|c|c|c|c|c|c|c|c|}
\hline
    \multirow{2}{*}{\makecell{Target}}  &
    \multirow{2}{*}{\makecell{Fault\\ Dia.}}  &  \multirow{2}{*} {\makecell{Motor\\Load}}  &  \multirow{2}{*}{\makecell{DNN \cite{sparseAE}}} &  \multirow{2}{*}{ \makecell{DANN \cite{dann_jmlr}}} &  \multirow{2}{*} {DTL \cite{lwen}} &  \multirow{2}{*} {DAFT \cite{wLu}} & \multicolumn{2}{c|} {Net2Net}  \\
\cline{8-9}
    & & & & & & &   \makecell{Without D. A.}  & \makecell{With D. A.}\\
\hline
\multirow{9}{*}{T1} & \multirow{3}{*}{\makecell{DE\\7 mil}} & 1hp &  92.6 & \textbf{99.7} &  93.2 &  93.5 &  95.3 & \textbf{99.7} \\
\cline{3-9} && 2hp &   89.4 & 90.6 & 92.4 & 90.4 & 97.6 & \textbf{98.3} \\
\cline{3-9} &&3hp  &   90.2 & 98.1 &  91.1 & 89.6 &   90.4 & \textbf{99.3} \\
\cline{2-9}
& \multirow{3}{*}{\makecell{DE\\14 mil}}&  1hp  &  72.9 & 33.1 & 71.0 & 71.7 & 84.9 & \textbf{90.2}\\
\cline{3-9} && 2hp  & 71.6 &   20.9 & 65.2 &  67.9 &  90.9 &  \textbf{97.9} \\
\cline{3-9} &&3hp  &  72.3 & 31.9 &  67.3 &  69.0 &  87.4 &  \textbf{95.5} \\
\cline{2-9}
& \multirow{3}{*}{\makecell{DE\\21 mil}} &  1hp  &  89.3 &  79.4 & 83.6 &  85.2 & 93.4 & \textbf{97.2}\\
\cline{3-9} && 2hp  &  90.9 & 52.2 &  87.8 &  88.0 &  90.6 &  \textbf{96.4} \\
\cline{3-9} && 3hp &  85.5 &  77.2 & 90.8 &  87.4 &  90.1 &  \textbf{97.7} \\
\hline
T2 & {\makecell{IMS\\Dataset}} &  26.6kN  &  83.81 &  85.63 & 82.94 &  81.59 & 87.59 & \textbf{91.66}\\
\hline
\rowcolor{Gray}\multicolumn{3}{|c|} {Standard Deviation} & 8.4 & 29.7 & 10.8 & 9.5 & 3.8 & 3.15\\
\hline
\multirow{9}{*}{\makecell{T1\\(10\% \\samples)}} & \multirow{3}{*}{\makecell{DE\\7 mil}} & 1hp &  62.1 & 90.1 &  83.1 &  80.8 &  89.8 & \textbf{96.0} \\
\cline{3-9} && 2hp &   52.4 & 76.3 & 81.3 & 80.9 & 85.7 & \textbf{92.9} \\
\cline{3-9} &&3hp  &   62.2 & 87.4 &  81.4 & 82.7 &   85.2 & \textbf{91.3} \\
\cline{2-9}
& \multirow{3}{*}{\makecell{DE\\14 mil}}&  1hp  &  58.4 & 35.3 & 61.8 & 60.9 & 84.8 & \textbf{91.5}\\
\cline{3-9} && 2hp  & 46.2 &  25.2 & 59.5 &  57.5 &  80.2 &  \textbf{90.8} \\
\cline{3-9} &&3hp  &  39.5 & 46.3 &  51.1 &  53.8 &  74.4 &  \textbf{87.0} \\
\cline{2-9}
& \multirow{3}{*}{\makecell{DE\\21 mil}} &  1hp  &  71.7 &  58.2 & 80.6 &  79.4 & 88.0 & \textbf{94.6}\\
\cline{3-9} && 2hp  &  63.0 & 42.7 &  83.0 &  82.7 &  91.7 &  \textbf{96.1} \\
\cline{3-9} && 3hp &  62.9 &  88.7 & 83.7 &  82.6 &  89.4 &  \textbf{91.3} \\
\hline
\makecell{T2\\(10\% \\samples)} & {\makecell{IMS\\Dataset}} &  26.6kN  &  59.2 &  81.4 & 80.0 &  78.2 & 85.6 & \textbf{91.6}\\
\hline
\rowcolor{Gray}\multicolumn{3}{|c|} {Standard Deviation} & 8.7 & 24.5 & 12.2 & 11.6 & 4.3 &2.02\\
\hline
\end{tabular}}
\label{Table: DE_DE}
\end{table*}
%=====================================================

\begin{table*}[ht]
\centering %
\caption{\textsc{\small  Accuracy on Paderborn University Dataset with different speed(N,M,F), load and radial force}} %
\resizebox{0.65\textheight}{!}{%
\begin{tabular}{ |c|c|c|c|c|c|c|c|}
\hline
    \multirow{2}{*}{\makecell{Target}}  &  \multirow{2}{*} {\makecell{Setting}}  &  \multirow{2}{*}{\makecell{DNN \cite{sparseAE}}} &  \multirow{2}{*}{ \makecell{DANN \cite{dann_jmlr}}} &  \multirow{2}{*} {DTL\cite{lwen}} &  \multirow{2}{*} {DAFT \cite{wLu}} & \multicolumn{2}{c|} {Net2Net}  \\
\cline{7-8}
    & & & & & &   \makecell{Without D. A.}  & \makecell{With D. A.}\\
\hline
 \multirow{4}{*}{T3} & L1  &  88.27 & 92.45 &  93.6 & 94.27 & 95.7 & \textbf{96.12} \\
\cline{2-8} & L2 &  78.73 & 90.23 & 85.6 & 88.93 & 91.2 & \textbf{93.6} \\
\cline{2-8} & L3  & 84 & 88.45 & 87.13 & 84.6 & 89.12 & \textbf{92.28} \\
\cline{2-8} & L4  & 78.4 & 87.6 & 88.4 & 87.13 &      93.12 & \textbf{93.6} \\
\hline
\multirow{3}{*}{T4}&  L1  &  89.53 & \textbf{95.5} & 94 &  94.47 & 90.5 & {90.1} \\
\cline{2-8} & L2  &  78.2 & 92.4 & 90.93 & 92 & 92.2 &      \textbf{94.31} \\
\cline{2-8} &L3 & 84.4 & 89.52 &  91.47 & 91 &  91.1 &   \textbf{95.2} \\
\cline{2-8} & L4  & 80.07 & 91 & 87.73 & 88.27 &    90.02 & \textbf{96.24} \\
\hline
\rowcolor{Gray}\multicolumn{2}{|c|} {Standard Deviation} & 4.5 & 2.5 & 3.1 & 3.5 & 2.06 & 2.05 \\
 \hline
 \multirow{4}{*}{\makecell{T3\\10\% \\samples}} & L1  &  58.6 & 82.3 & 73.5 & 80.0 & 88.0 & \textbf{92.3} \\
\cline{2-8} & L2 &  56.3 & 90.2 & 71.4 & 81.0 & 86.5 & \textbf{93.0} \\
\cline{2-8} & L3  & 70.2 & 58.2 & 84.1 & 85.3 & \textbf{90.0} & 89.0 \\
\cline{2-8} & L4  & 64.4 & 89.0 & 71.0 & 77.3 & 78.2 & \textbf{85.0} \\
\hline
\multirow{3}{*}{\makecell{T4\\10\% \\samples}}&  L1  &  65.3 & 55.0 & 76.0 & 76.0 & 78.0 & \textbf{84.5} \\
\cline{2-8} & L2  &  58.5 & 48.1 & 76.0 & 80.0 & 85.2 & \textbf{90.0} \\
\cline{2-8} &L3 & 64.8 & 73.2 & 78.2 & 88.0 & 84.2 & \textbf{92.0} \\
\cline{2-8} & L4  & 60.0 & 90.0 & 79.3 & 83.4 & 87.0 & \textbf{90.0} \\
\hline
\rowcolor{Gray}\multicolumn{2}{|c|} {Standard Deviation} & 6.4 & 17.3 & 4.4 & 4.0 & 4.4 & \textbf{3.2} \\ 
 \hline
\end{tabular}}
\label{Table: paderborn}
\end{table*}
%****************************************************

\begin{figure*}[!ht]
\centering
\hspace{-1cm}
\includegraphics[width=16.0cm]{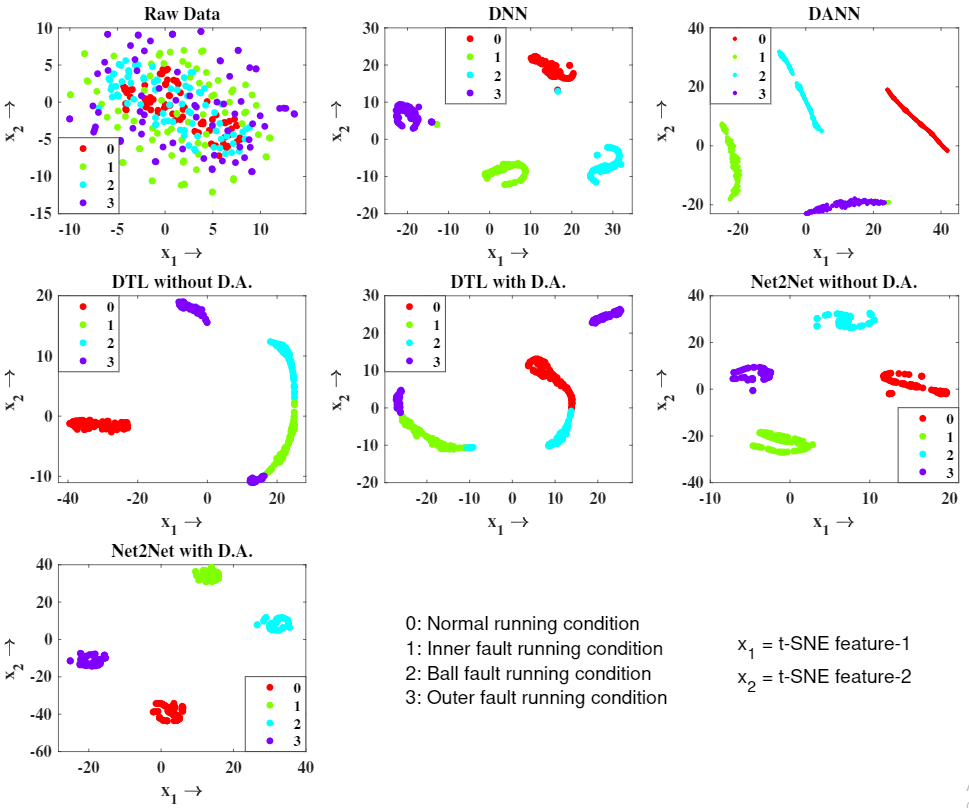}
\caption{Feature visualization for CWRU DE Fault Data (with 7 mil fault diameter and 1 hp motor load) using t-SNE.}
\label{Fig:t-SNE}
\end{figure*}

\subsection{Segmentation and Evaluation Scheme}
The recorded data files have the time-series signals and there is a huge number of sample points (at least 121265 points) in each class which may not be directly suitable for training the DNN. Therefore source and target datasets are prepared by segmenting the recorded samples with the segment length of approx $100$ data points for the CWRU/IMS dataset and $400$ data points for the Paderborn University dataset. For example, if a time-series signal contains 121000 points, it can be converted into $1210\times100$ data samples per class.

Effectiveness of the domain adaptation of the proposed algorithm is studied in two cases:
\subsubsection{Case-I} CWRU dataset + IMS dataset: The source and the target dataset represents four classes: Healthy/Normal (N), Inner race (IR), Ball (B) \& Outer race (OR) as provided in Table \ref{tab:dataD1}.
\begin{enumerate}[i)]
    \item \textbf{Source Data:} Source dataset is created from \textit{12k Hz Drive End (DE)} fault with \textit{fault dia. = $7$ mil \& load = $0$ hp}. With the segment length of $100$ data points, $1210\times100$ samples per class are created. Therefore, source data has the size of $4840\times100$.
    \item \textbf{Target-1:} The target-1 (T1) is prepared using the CWRU dataset for the \textit{12k Hz DE} fault with 7, 14 \& 21 mil fault diameter and at 1, 2 \& 3 hp motor load. For each cases, 40,000 samples points from the time-series signal are used to create a dataset of $400\times100$ per class.
    \item \textbf{Target-2:} The target-2 (T2) is prepared using \textit{IMS dataset} recorded at 26.6 kN motor load. 40,000 data points are taken to create a dataset of $400\times100$ per class.
\end{enumerate}

\subsubsection{Case-II} Paderborn university dataset: The source and the target dataset represents three classes: Healthy/Normal(N), Outer Race (OR) \& Inner Race (IR).
\begin{enumerate}[i)]
    \item \textbf{Source Data:} Source dataset is created using artificially damaged fault dataset with the extent of damage for OR \& IR fault = 1. Each measurement file contains approx 256001 sample points as a time-series signal.  With the segment length of $400$ data points, $10000\times400$ samples per class created using 20 measurement data files from each class as mentioned in Table \ref{tab:dataD2}.
    \item \textbf{Target-3 \& 4:} target data (target-3 (T3) \& target-4 (T4)) are selected from real damaged fault dataset as summarized in table \ref{tab:dataD2}. Samples for both the target datasets (T3 \& T4) are considered under four load settings L1, L2, L3, \& L4. $500\times400$ samples per class is created using one measurement file from each class.
\end{enumerate}
The selection of the source and the target datasets as described above has also been pictorially shown in Fig. \ref{Fig: flow diagram}. 

% \begin{enumerate}
%     \item \textbf{Target-2:} Target-2 data consists of normal and faulty data from CWRU, \textit{12k Hz Fan End (FE)} fault with 7, 14 \& 21 mil fault diameter and at 1, 2 \& 3 hp motor load. For each cases, 40,000 samples points are used to create a dataset of $400\times100$ per class.
% \end{enumerate}

\subsection{Training and Results}  For the demonstration of the efficacy of the proposed method, first, the teacher model (DNN with hidden layers: $70-30-20$) is trained on the source data using the algorithm \ref{algo:Te_DNN}. The hyper-parameter of the DNN for its training are takes as: regularization parameter ($\lambda$) = 0.05, sparsity parameter ($\rho$) = 0.1, weight penalizing deviation ($\beta$) = 0.8, method of parameter optimization = 'lbfgs'.

This model is used to train DNN with new architecture (here we have selected the new architecture as $70-50-30-20$) suitable for the machine running on different operating conditions. 
The target datasets from each cases as described in the Tables \ref{tab:dataD1} \& \ref{tab:dataD2} are normalized and then split into train-test datasets using five fold cross-validation sampling technique. Now, the new model is trained using the algorithm-\ref{algo:net2net} on the training datasets from the target-1, 2, 3 \& 4.  Parameter optimization of the new model is done using back propagation with softmax as classifier. The network is trained for 50 iteration with adaptive learning rate. 
Classification accuracies on the test datasets from each case are generated and presented in the Tables \ref{Table: DE_DE} and \ref{Table: paderborn}.

To show the effectiveness of the proposed method, we have compared the results with the most advanced and similar methodologies of the fault diagnosis of rotating machines.  As reported in literature review the selected algorithms are deep neural network (DNN) \cite{sparseAE}, domain adversarial neural network (DANN) \cite{dann_jmlr}, deep transfer learning (DTL) with classification loss and MMD term minimization \cite{lwen} and Deep Model Based Domain Adaptation \cite{wLu}. All these models are also trained using the same dataset. The architecture of the DNN and the DTL is kept the same as the new model (student net): $70-50-30-20$.

DNN with the new architecture has been trained from scratch on the target datasets: T1, T2, T3, and T4. DANN has been trained using labeled source data and the unlabeled target data \cite{dann_jmlr}. DTL with the new architecture is pre-trained on unlabeled source data and fine-tuned on the target data based using the method in \cite{lwen} and \cite{wLu}. The High-level feature output of the deep learning algorithm is used to trained the softmax classifier (SC) and classification accuracies are presented in Tables \ref{Table: DE_DE} and \ref{Table: paderborn}. Average training time under the same computational condition for each algorithm has been compared in Table \ref{tab:timeComp}.  The h-level feature visualization of all the methods for the CWRU DE fault dataset (T1, 7 mils, 1 hp) has been presented in Fig. \ref{Fig:t-SNE}.

\begin{table*}[!ht]
\centering %
\caption{\textsc{\footnotesize Average Training Time ($Sec.$) on the Same Machine under identical condition}} %
% \resizebox{0.36\textheight}{!}{%
\begin{tabular}{ |c|c|c|c|c|c|c|}
\hline
     \multirow{2}{*} {\makecell{Target\\Data Name}}  &  \multirow{2}{*}{\makecell{DNN \\ \cite{sparseAE} }} &  \multirow{2}{*}{ \makecell{DANN\\ \cite{dann_jmlr}}}&  \multirow{2}{*} {DTL\cite{lwen}} &  \multirow{2}{*} {DAFT \cite{wLu}} & \multicolumn{2}{c|} {Net2Net (Proposed)}  \\
\cline{6-7}
     & & & & &   \makecell{Without D. A.}  & \makecell{\; With D. A. \; }\\
\hline
      Target-1 & 165.84 & 2476.26 & {238.6} & 206.8 & \textbf{90.4} & \textbf{55.0}\\
      \hline
      Target-2 & 160.96 & 2490.4 & {278.1} & 260.9 & \textbf{80.2} & \textbf{61.8}\\
      \hline
      Target-3 & 156.96 & 2490.4 & {278.1} & 260.9 & \textbf{80.2} & \textbf{61.8} \\
      \hline
      Target-4 & 426.96 & 4490.4 & {978.1} & 960.9 & \textbf{215.2} & \textbf{191.8} \\
\hline
\end{tabular}
\label{tab:timeComp}
\end{table*}

%=============================================================

\subsection{Discussion} Based on the results shown in the Tables \ref{Table: DE_DE} \& \ref{Table: paderborn} and the feature visualization in the Fig. \ref{Fig:t-SNE}, following points can observed.
\begin{enumerate}[i)]
     \item The proposed method (Net2Net with D.A) performs better for all operating conditions as compared to the state-of-the-art methods. Here, DNN fails to perform on the par because it is trained on the small target data (insufficient training data) from scratch. Similarly, DTL with D.A. overfits due to an insufficient amount of the target data. In the case of the proposed method,  the knowledge of the source data is transferred via function preserving principle to the new network with a suitable architecture. The transformed model is almost ready for the new data pattern. Further, the transformed model is fine-tuned using the target data to minimize the classification loss and the distribution discrepancy. Therefore, the proposed method performs better even if a very less amount of the target data is available.  
     \item The comparison of the standard deviation (S.D)  over the various operating conditions shows the stability of the performance. It can be observed that the results by the Net2Net with D.A are more stable with the variations in the operating conditions. Results of the DANN are the most unstable because DANN tries to adapt with an unlabelled target dataset which is very small in number.
     \item For the Paderborn University dataset, source data is taken from the machine with artificially damage faults and the target datasets are from real damaged (run to failure case). From \ref{Table: paderborn}, it can be observed that the results with Net2Net are stable even under wide variations of the operating conditions and the extent of the damage.
    %  \item DTL is pre-trained on the source dataset and fine-tuned using the small number of the target dataset. The performance of transfer learning is very poor due to the small number of training samples and the domain shift from source to target. Again, if DTL is fine-tuned with D.A. \cite{wLu}, it performs better.
     \item The Fig. \ref{Fig:t-SNE} shows h-level feature visualization using the t-SNE. It can be observed that the h-level features from Net2Net with D.A are more clearly separated from each other as compared to the state-of-the-art methods.
     \item The performance of the proposed method under a reduced number of available labelled target samples (10\% of the samples) is also presented in Table \ref{Table: DE_DE} and \ref{Table: paderborn}. For the reduced number of the target samples, the effect of the domain adaptation (in Eq. (9)) is reduced, therefore, the reduction in the performance. However, the performance of the proposed method is still better than the other state-of-the-art methods because of the knowledge transfer through network transformation.
     \item If the number of training samples or the number of training iterations is reduced, the training time will be reduced, but it may result in overfitting of the network training leading to poor accuracy. On the other hand, if we wish to further improve the accuracy, more samples will be required, but, this will increase the computational time.
\end{enumerate}

\subsection{Complexity of the Proposed Method:} The training process of the proposed algorithm includes (i) network transformation (initialization using function preserving principle) and (ii) the fine-tuning to obtain optimal parameters by solving the minimization objective in Eq. (\ref{Eq:MinArg}).
The network transformation includes the conversion, insertion, and/or removal elements from parameter matrices. Therefore, the time complexity of the whole training process is mainly contributed by the fine-tuning process to get the optimality of Eq. (\ref{Eq:MinArg}) which is similar to training the deep neural network using the back propagation algorithm.
Let $n$ denotes the number of data points, $d$ dimensionality of the input data, $i,\,j,\,k$ are the number of nodes in the hidden layers in the network, $c$ be the number output class, $N$ be the number of iterations required for the training. The time complexity analysis using $O(.)$ \cite{tcomplexity} for the process of fine-tuning can be expressed as $O(N*n*(d*i+i*j+j*k+k*c))$

Since the fine-tuning requires a very less amount of labelled target data and only a few iterations, the training process takes very little time. The time complexity of each method listed is different because the training process in each method is different. However, assuming the same task of the fault diagnosis to be performed, the time comparison in Table V gives useful information about the quick adaptation of the proposed method under the variable operating conditions.

\section{Conclusions} \label{sec:conclusions}
In this paper, we have proposed a quick learning methodology based on Net2Net transformation followed by a fine-tuning to minimize classification loss and domain discrepancy. The proposed method is very effective for intelligent fault diagnosis under the variable operating conditions of industrial rotating machines. The fine-tuning process requires a very less amount of target data and a few number of iterations for fine-tuning, therefore, the method is capable for quick adaptation to the change in the operating condition. Also, the performance of the proposed method is stable under the variable operating conditions compared to the state-of-the-art methods. Therefore, the proposed method is very useful for continuous monitoring of the industrial machines.

The proposed method of network transformation has a broad scope in the future for developing lifelong learning systems. This work can be extended to develop a network optimization algorithm capable of adjusting its architecture depending upon the change in the operating conditions.

\bibliographystyle{IEEEtran.bst}
\bibliography{Ref.bib}

\end{document}